\documentclass[a4paper,conference]{IEEEtran}
\IEEEoverridecommandlockouts
\usepackage{cite}
\usepackage{amsmath,amssymb,amsfonts}
\usepackage{algorithmic}
\usepackage{graphicx}
\usepackage{textcomp}
\usepackage{xcolor}
\usepackage{amsmath}
\usepackage{algorithm}
\usepackage{url}
\usepackage{bm}
\usepackage{amsmath}
\usepackage{amssymb}
\def\BibTeX{{\rm B\kern-.05em{\sc i\kern-.025em b}\kern-.08em
    T\kern-.1667em\lower.7ex\hbox{E}\kern-.125emX}}
\begin{document}

\title{RSAC: Regularized Subspace Approximation Classifier for Lightweight Continuous Learning\\
}

\author{\IEEEauthorblockN{Chih-Hsing Ho}
\IEEEauthorblockA{\textit{Department of Electrical Engineering and Computer Science Honor Program} \\
\textit{National Chiao Tung University}\\
Hsinchu, Taiwan \\
bill86416.eecs04@nctu.edu.tw}
\and
\IEEEauthorblockN{Shang-Ho (Lawrence) Tsai}
\IEEEauthorblockA{\textit{Department of Electrical Engineering} \\
\textit{National Chiao Tung University}\\
Hsinchu, Taiwan \\
shanghot@alumni.usc.edu}
}

\maketitle

\begin{abstract}
Continuous learning seeks to perform the  learning on the data that arrives from time to time. While prior works have demonstrated several possible solutions, these approaches require excessive training time as well as memory usage. This is impractical for applications where time and storage are constrained, such as edge computing. In this work, a novel training algorithm, regularized subspace approximation classifier (RSAC), is proposed to achieve lightweight continuous learning. RSAC contains a feature reduction module and classifier module with regularization. Extensive experiments show that RSAC is more efficient than prior continuous learning works and outperforms these works on various experimental settings.
\end{abstract}


\section{Introduction}
Deep networks have enabled significant advances over the last decade in many machine learning tasks, such as image classification~\cite{alexnet,resnet,vgg} and object recognition~\cite{rcnn,maskrcnn,yolo,voxnet,ranet,capsulevos}.
The success is especially significant under the setting of supervised learning, where the entire labeled dataset is provided to train the deep network to complete the assigned task (i.e. image classification). Despite the success of supervised learning, its success is often achieved on the presumption that the tasks are assigned all at once. This is not a realistic learning procedure as human.
As a realistic learner, human possess the ability to continually grow the knowledge throughout the lifespan by solving different tasks. The supervisions of those tasks from different time span assist the establishment of human capability~\cite{Tani16,bremner11}. While human benefits from the continuous supervision and the shift of tasks from the environment, this is not the case for deep network, which fails on solving the old tasks when a new task is learned~\cite{Goodfellow13,icarl,survey}.
Such phenomenon has been referred as \textit{catastrophic forgetting}~\cite{catforget,catforgetinconnect,measure} and its potential solutions are discussed in the literature of continuous learning ~\cite{continualsur,survey,scenarios,Goodfellow13}. 

Continuous learning seeks to robustify the knowledge previously learned by deep network. By consolidating the knowledge, the network can be adopted in the scenario where data continuously streams in and achieves good performance on new coming tasks without forgetting the old ones. The mainstream solutions of performing continuous learning are knowledge consolidation~\cite{lwf,ewc,si,mas}, network expansion~\cite{Jeongtae17,RusuRDSKKPH16,xiao2014,pathnet}
and memory rehearsal~\cite{dgr,icarl,pseudo}. 
While prior works provide several possibilities to avoid catastrophic forgetting, the training time is usually exhaustive~\cite{latent,survey}
and thus hinders the application of continuous learning in real world scenario, where the model not only has to perform well in the new tasks, but also has to complete the learning within a short period to keep track of the new coming data. Moreover, for the applications on edge computing, the memory usage is also another concern. In this paper, we consider these constraints and refer the problem as lightweight continuous learning (LCL). 

To address this problem, a novel algorithm,
\textbf{regularized subspace approximation classification (RSAC)}, is proposed which contains a transformation module and a classifier module.
For the transformation module, a lightweight feature reduction algorithm is introduced. This is inspired by prior works~\cite{saab,saak} that discard the use of backpropagation in deep network and substitute with a set of explainable modules. For the classifier, the quadratic discriminant classifier (QDC)~\cite{qdc} is used with the proposed regularization. By combining these 2 modules, RSAC expedites the training under low memory usage without sacrificing the performance.


\begin{figure}
    \centering
    \includegraphics[width=\linewidth]{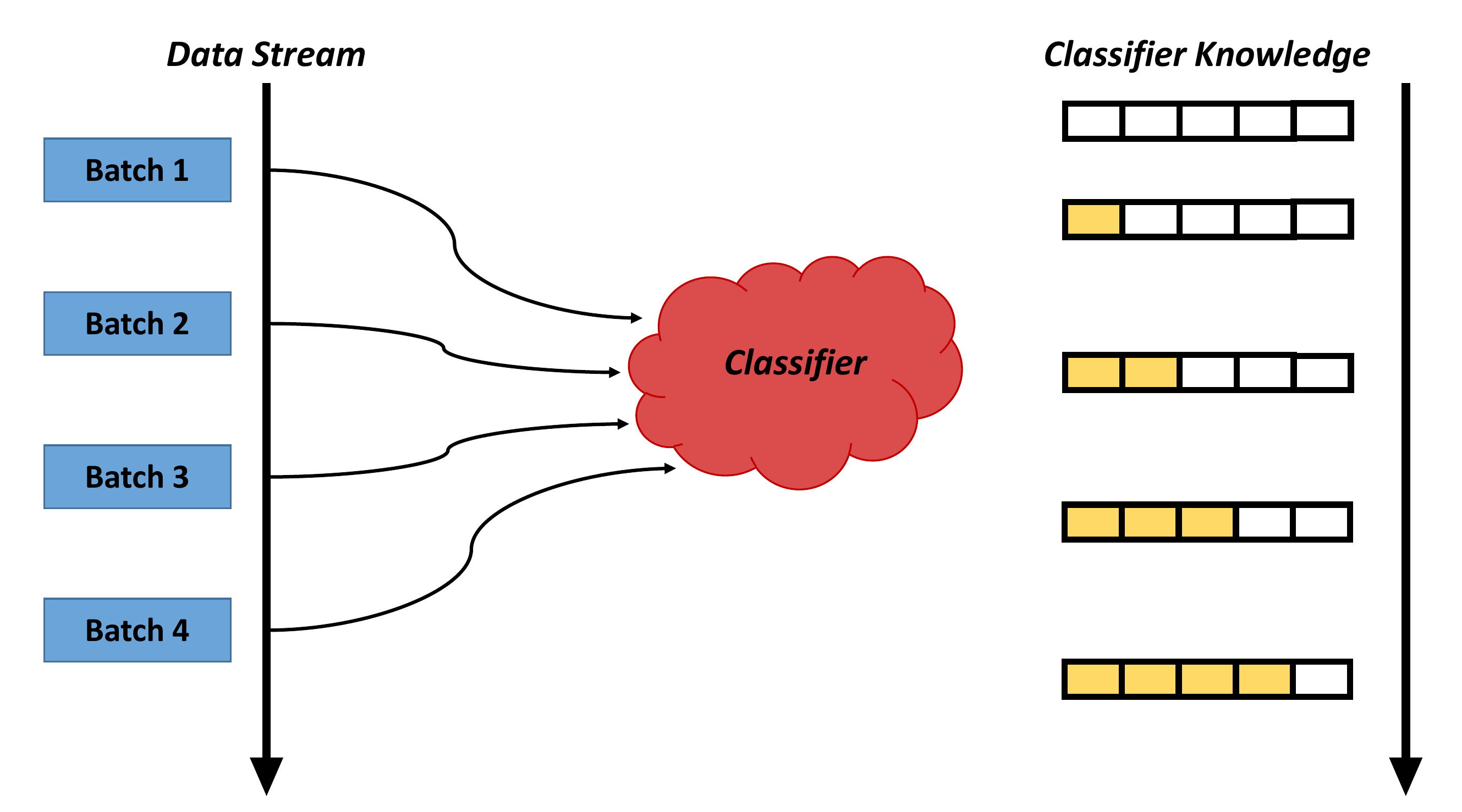}
    \caption{A lightweight continuous learning algorithm is designed to perform the learning as data streams in under the constraint that both training and inference time is short and memory usage is low.}
    \label{fig:intro_fig}
\end{figure}

In summary, the contribution of this work is 3 folds. First, the current limitations of continuous learning methods, including large memory usage and long training time, are discussed. The problem of lightweight continuous learning (LCL) is then formulated and investigated.
Second, we proposed a novel algorithm, regularized subspace approximation classifier (RSAC), for LCL problem.
Finally, extensive experiments demonstrate that RSAC achieves state-of-the-art performance under different continuous learning settings with large computation time improvement. The proposed model provides a promising solution to the real world continuous learning applications under various practical constraints.

\section{Related work}
In this section, the limitations of previous continuous learning works are discussed and the prior lightweight transformation and classification algorithms, which this work is inspired of, are reviewed.

\subsection{Continuous learning}

The study of continuous learning seeks to mitigate the catastrophic forgetting phenomenon~\cite{MCCLOSKEY1989109,ewc,survey}, where the classifier forgets the knowledge previously established after training on new data. The proposed approaches can be mainly categorized into weight consolidation~\cite{knowledge_distill,lwf}, architecture expansion~\cite{Jeongtae17,RusuRDSKKPH16,xiao2014,pathnet} and memory rehearsal~\cite{Jeongtae17,RusuRDSKKPH16}. Their infeasibility on applying to lightweight continuous learning (LCL) scenario is discussed.

Weight consolidation based methods~\cite{knowledge_distill} impose constraint to avoid dramatic shift on the learned weights. The constraint encourages similar output activation between the use of the current weights and the updated weights. This can be implemented through knowledge distillation loss~\cite{lwf}, L2 regularization~\cite{JungJJK16} or updating the weight with smaller learning rate~\cite{donahue14}. In addition, these constraints can be explicitly applied to those weights that are important to specific task~\cite{ewc,si,mas}. However, these methods often suffer from insufficient learning capacity as the flexibility is restricted by the regularization~\cite{survey} imposed for consolidating the old knowledge. Moreover, methods like LwF~\cite{lwf} are not applicable for LCL scenario as the training time increases proportionally with the number of tasks.

Expanding the architecture dynamically is another type of solutions for continuous learning and breaks the inflexibility limitation of weight consolidation based methods. The expansion can be performed by allocating a new subnetwork~\cite{Jeongtae17,RusuRDSKKPH16,Adanet,xiao2014} or inserting neurons in a hierarchical manner~\cite{xiao2014}. Aside from adding completely new modules, useful neurons from trained feature extractor can be selected by leveraging dynamic path~\cite{pathnet} or gating functions~\cite{Nicolas18}. In addition, to prevent overfitting of the incrementally larger network, \cite{zhou12b} merges neurons with similar response for downstream training.
However, these approaches are difficult to scale up in general when new coming tasks increase dramatically and thus are not applicable to the LCL scenario. 

Rehearsal based approaches~\cite{Tyler18,Lee19,icarl,Francisco18,fearnet} have demonstrated recent success on continuous learning by training on few examples from the previous batches, which can be sampled from the storage with limit memory~\cite{icarl,Francisco18} or from the deep generators~\cite{fearnet,Nitin17,dgr,rtf} (i.e. autoencoder~\cite{autoencoder} and generative adversarial network\cite{gan}). However, explicitly storing past examples requires extra memory usage and the use of deep generators needs additional training. Again, all these prior works do not meet the requirement of LCL, which emphasizes on short training time, memory efficient and fast inference.

\subsection{Lightweight transformation and classification}

Despite the recent success of deep networks, it requires large computation resources and training time, which hinders its real world applications especially on edge computing.
These drawbacks are addressed with subspace approximation with augmented kernels (Saak)~\cite{saak} and its variant, subspace approximation with adjusted bias (Saab)\cite{saab}. Saak transform is an interpretable one-pass feedforward network based on truncated Karhunen-Loève Transform (KLT)~\cite{klt}, or PCA~\cite{pca}, that transforms the input domain (i.e. spatial domain for images) into a latent domain to obtain its associated latent representation.
In order to approximate functions with higher complexity that maps between the input and the latent representation, cascaded Saak transformations are used together with the ReLu function in between individual Saak transformation. However, the use of ReLu function sacrifices partial information as the value below zero are truncated.
To minimize the loss of information, a negative counterpart of the kernel vectors from truncated KLT is introduced, such that all the information will be preserved after projecting on the kernel vectors from truncated KLT and its negative counterpart. 

Despite the advantages of Saak transform, the size of latent representation will grow exponentially with more cascade of Saak, due to the use of additional kernel vectors. To overcome this drawback, subspace approximation with adjusted bias (Saab)~\cite{saab} transformation adds a computed bias term on each of the projection on KLT kernel vectors. Thus, Saab not only inherits the advantages of Saak transform, but further improves the memory efficiency. In addition, Saak transform and Saab transform have demonstrated competitive results compared to deep neural network on different domains, including image classification~\cite{saak,saab,pixelhop,pixelhopplus}, 3D object classification~\cite{pointhop} and texture analysis~\cite{hssc}. Moreover, they can be applied in semi-supervised learning~\cite{semisupervised} and image compression~\cite{imagecompression} and demonstrate robustness toward adversarial attacks~\cite{robustsaak}. Inspired by the characteristic of these transformations in terms of memory efficient, lightweight computation and success in multiple domains, we explore whether the same benefits ensue in the scenario of lightweight continuous learning.

\begin{figure*}
    \centering
    \includegraphics[width=0.9\linewidth]{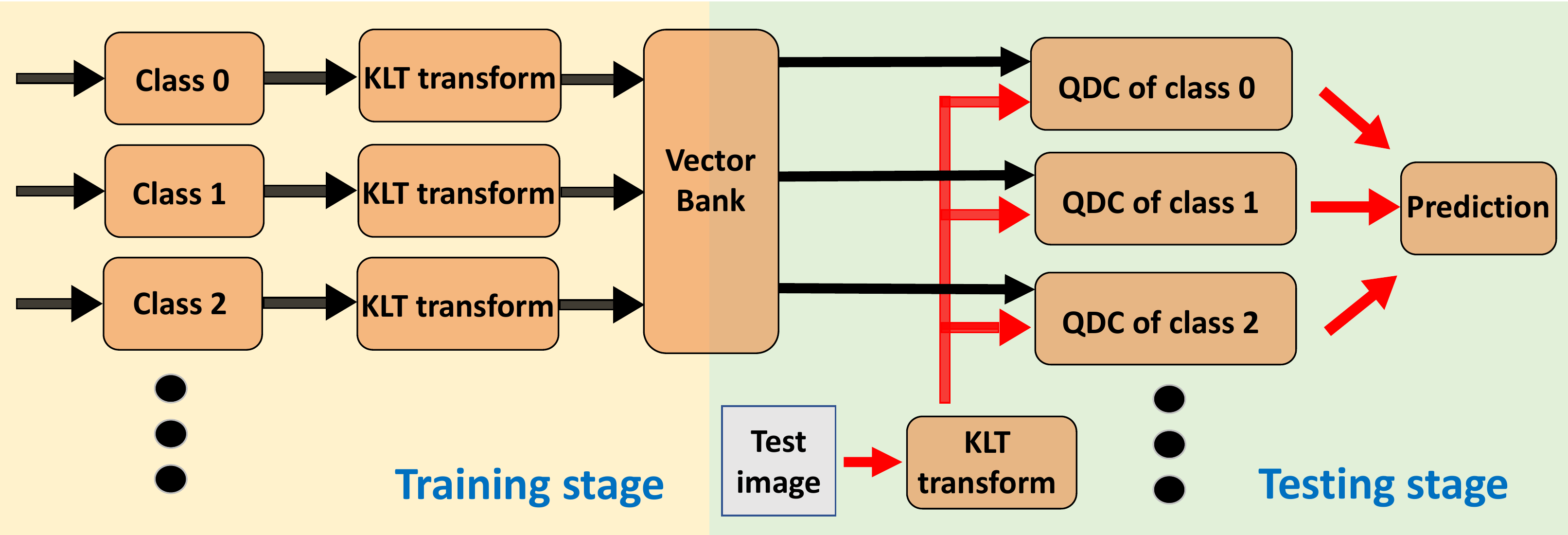}
   \caption{
    The flow chart of the proposed RSAC architecture for lightweight continuous learning. During training stage, the eigenvectors of the KLT transform will be stored in the vector bank after learning. During inference stage, the dimension of data from each class will be reduced through KLT transform. The reduced feature will then be classified with the QDC classifier.}
    \label{fig:main_fig}
\end{figure*}

\section{Method}
 
In this section, the proposed algorithm regularized subspace approximation classifier (RSAC) for lightweight continuous learning (LCL) is introduced.\

\subsection{Preliminary formulation}\label{sec:setting}
Given a labeled dataset $\mathcal{D}_{full}=\{x_i, y_i\}_{i=1}^{N}$, where $x_i\in\mathcal{R}^d$ is an image, $\mathcal{X}_{full}=\{x_i\}_{i=1}^{N}$ is a full image set, $y_i \in \mathcal{Y}_{full}=\{1 \dots C\}$ is a full label set and $C$ is the number of classes.
A classifier $F$ is trained to solve the assigned task originated from the dataset $\mathcal{D}_{full}$. Let $\mathcal{T}$ be the set of tasks to be solved by the classifier $F$. For a classification problem on image set $\mathcal{X}=\{x_i\}$ and label set $\mathcal{Y}=\{y_i\}$, the task $t(\mathcal{X}, \mathcal{Y})$ is defined to classify an image $x$ in $\mathcal{X} \subseteq \mathcal{X}_{full}$ to the target label set $\mathcal{Y}  \subseteq \mathcal{Y}_{full}$. In other words, the classifier aims to maximize the posterior class probability
\begin{equation}
    P_{Y|X}(c|x) = \frac{P_{X|Y}(x|c)P_Y(c)}{P(X)},
    \label{eq:bayes}
\end{equation}
which can be expressed with the likelihood $P_{X|Y}(x|c)$ and class prior $P_Y(c)$ with Bayes rule~\cite{bayes}.

For the supervised learning scenario, the entire labeled dataset are used during the training procedure, where $\mathcal{T}=\{t_{1}(\mathcal{X}_{full}, \mathcal{Y}_{full})\}$ as there is only a single task.
While for continuous learning with $M$ different tasks, the task set is defined as \begin{equation}
\mathcal{T}=\{t_j(\mathcal{X}_j,\mathcal{Y}_j)\}_{j=1}^M,
\label{eq:task_def}
\end{equation}
and satisfies the constraint that 
\begin{equation}
\mathcal{X}_i\cap\mathcal{X}_j=\O, \quad i\neq j, \quad \bigcup\limits_{j=1}^{M} \mathcal{X}_j=\mathcal{X}_{full}.
\end{equation}
Correspondingly, the label set has to satisfy $\mathcal{Y}_i\cap\mathcal{Y}_j=\O$ and $ \bigcup\limits_{j=1}^{M} \mathcal{Y}_j=\mathcal{Y}_{full}$.  
Note that the general definition of task set $\mathcal{T}$ can be applied to supervised learning by setting $M=1$.

\subsection{Deep continuous learning}
While the classifier $F$ can be implemented in many different ways, one of the common manners is to leveraged the deep neural network model. Under the general definition of task set $\mathcal{T}$, the classifier can be formulated as $F:\mathcal{R}^d\rightarrow\mathcal{R}^{C}$.  This is implemented by the combination of a feature extractor $f_\theta(x)$ of parameters $\theta$ and a softmax regression layer
that predicts the posterior class probability as
\begin{equation}
    F_c = P_{Y|X}(c|x) = \frac{e^{w_c^T f_\theta(x)}}
    {\sum_{k=1}^C e^{w_k^T f_\theta(x)}}, 
    \label{eq:deep_cls}
\end{equation}
where $F_c$ denotes the $c^{th}$ entry of classifier $F$, $w_c$ is the vector of classification parameters of
class $c$. However, the learning of $\theta$ is often not transparent and fails to yield an explainable result~\cite{explainai}. Moreover, the learning of $\theta$ requires the backpropagation procedure, which is computation expensive and relies on great amount of computing resources. All these drawbacks impede the advance of lightweight continuous learning (LCL) in real world applications.

\subsection{Lightweight continuous learning}
To avoid high computation cost, regularized subspace approximation classifier (RSAC) is proposed in this section as a solution for lightweight continuous learning (LCL) and contains 2 modules, including a feature reduction module and a classifier module. 

\subsubsection{Feature reduction module}\label{sec:saab_plus}

Feature reduction is a critical stage in entire LCL pipeline, as simply taking raw image $x\in\mathcal{R}^d$ as input to the classifier will lead to high computational cost. For deep continuous learning, the feature reduction is performed by the feature extractor $f$ of (\ref{eq:deep_cls}), while the KLT transformation is leveraged in lightweight continuous learning, inspired by the lightweight Saab transformation used in supervised learning.
KLT is established on 
the covariance matrix $\Sigma_c$ of class $c$ computed by
\begin{equation}
\mu_c = \frac{1}{N_c}\sum_{j=1}^{N_c}x_j,
\label{eq:mu}
\end{equation}
and
\begin{align}
\Sigma_c &= \frac{1}{N_c}\sum_{j=1}^{N_c}(x_j-\mu_c)(x_j-\mu_c)^T \nonumber\\
&= Q_c\Lambda_cQ_c^T, 
\label{eq:sigma}
\end{align}
where $N_c$ is the number of data belongs to class $c$, $\Lambda_c$ is a diagonal matrix with the eigenvalue $\sigma_c^j$ as the $j^{th}$ entry and $Q_c$ is a $d\times d$ orthonormal matrix, where the $j^{th}$ column vector is the eigenvector $q_c^j$.
The feature reduction is then implemented by projecting on the top $k$ eigenvectors in $Q_c$ correspond to the $k$ largest eigenvalues in $\Lambda_c$, which are selected as
\begin{equation}
    \frac{\sum_{j=1}^{k} \sigma^j_c}{\sum_{j=1}^{d} \sigma^j_c} \geq t,
    \label{eq:power_thres}
\end{equation}
where $t$ is a power threshold to guarantee that sufficient information is preserved. Let $\hat{Q}_c$ be a $d\times k$ matrix, where the column vectors are the concatenation of $k$ selected eigenvectors, and $\hat{\Lambda}_c$ is a $k\times k$ diagonal matrix with the top $k$ largest eigenvalues in $\Lambda_c$. The resulting latent representation after projection is denoted as $f(x)=\hat{Q}_c^Tx$. Note that unlike the feature extractor $f$ in (\ref{eq:deep_cls}), no parameter is needed to be learned in feature reduction module. Moreover, the power threshold $t$ can be used to control the number of eigenvectors stored in the vector bank with limited memory buffer, as shown in the training stage of Fig \ref{fig:main_fig}. With such projection scheme, the input image $x$ can be represented with $f(x)\in \mathcal{R}^k$.


\subsubsection{Classifier module}
To discard the learning of numerous parameters in deep neural network, the relationship of (\ref{eq:deep_cls}) and (\ref{eq:bayes}) is investigated. We then note that maximizing the class posterior probabilities of (\ref{eq:bayes}) can be reformulated as 
\begin{equation}
    \max P_{Y|X}(c|x) = \max P_{X|Y}(x|c)P_Y(c),
    \label{eq:max_bayes_drop}
\end{equation}
by dropping the denominator $P(x)$ in (\ref{eq:bayes}), since the distribution of the input data $X$ is independent of the learning.
In general, the class conditional distribution $P_{X|Y}(x|c)$ can be 
modeled with any function in the exponential family distribution~\cite{Sundberg2011}, i.e. 
\begin{equation}
    P_{X|Y}(x|y) = q(x)e^{<w_y,v(x)>-\phi(w_y)}
    \label{eq:exponential_family}
\end{equation}
and 
\begin{equation}
    P_{Y}(y) = \frac{e^{\phi(w_y)}}{\sum_{i=1}^C e^{\phi(w_i)}},
    \label{eq:exponential}
\end{equation}
where $w_y$ is a canonical parameter, $v(x)$ is a sufficient statistic, $\phi(w_y)$ is a cumulant function and $q(x)$ is a
underlying measure~\cite{Sundberg2011}. 
While, in principle, any function in the exponential family can be leveraged, a simple Guassian distribution is considered in this work. The likelihood that  operates on the latent representation $f(x)\in\mathcal{R}^k$ from feature module
is then modeled with 
\begin{align}
P_{X|Y}(f(x)|c) &= \mathcal{G}(f(x);\hat{\mu}_c,\hat{\Sigma}_c) \nonumber\\
&=\frac{1}{(2\pi)^{d/2}|\hat{\Sigma}_c|}e^{-\frac{1}{2}(f(x)-\hat{\mu}_c)^T\hat{\Sigma}_i^{-1}(f(x)-\hat{\mu}_c)}.
\label{eq:pxy}
\end{align}
Similarly, the class prior can be computed as 
\begin{equation}
P_{Y}(c) = \frac{N_c}{\sum_{j=1}^C N_j}.
\label{eq:py}
\end{equation}
Note that the computation of $\hat{\mu}_c$ and $\hat{\Sigma}_c$ of class $c$ is similar to (\ref{eq:mu}) and (\ref{eq:sigma}), but operates on the latent representation $f(x)$. 

As shown in the inference stage of Fig~\ref{fig:main_fig}, an input example $x$ is mapped to the latent representation $f(x)$ with the feature reduction module and then classified by computing the maximum a posteriori as  
\begin{equation}
\mathop{\arg\max}_{c}  P_{Y|X}(c|f(x)) = \mathop{\arg\max}_{c}  P_{X|Y}(f(x)|c)P_Y(c).
\label{eq:map}
\end{equation}
By substituting the modeling of (\ref{eq:pxy}) and (\ref{eq:py}) into (\ref{eq:map}), it can be re-written in log scale as
{\scriptsize
\begin{align}
&\mathop{\arg\max}_{c}  ln(P_{Y|X}(c|f(x)))  \nonumber\\ 
&=
\mathop{\arg\max}_{c}  ln(P_{X|Y}(f(x)|c)) + ln(P_Y(c)) \nonumber\\
&=
\mathop{\arg\max}_{i}  ln(\frac{1}{|\hat{\Sigma}_c|})-\frac{1}{2}(f(x)-\hat{\mu}_c)^T\hat{\Sigma}_c^{-1}(f(x)-\hat{\mu}_c)+ln(\frac{N_c}{\sum_{j=1}^C N_j})
\label{eq:map_log}
\end{align}}
This is referred as quadratic discriminant classifier (QDC) in the following.
\begin{algorithm}[t]
\caption{Pseudocode of RSAC}
\label{alg:main}
\begin{algorithmic}[1]
\STATE{EigList:= []}
\WHILE {Training}
\WHILE {Data $\mathcal{X}_c$ from new class $c$ stream in}
\STATE{Apply feature reduction module in Sec.~\ref{sec:saab_plus}.}
\STATE{Append the $k$ eigenvectors of class $c$ selected from (\ref{eq:power_thres}) to EigList.}
\ENDWHILE
\ENDWHILE
\WHILE {Testing}
\FOR{all test images $x$}
\STATE{Apply feature reduction module in Sec.~\ref{sec:saab_plus}.}
\STATE{Perform QDC of (\ref{eq:map_log}) with regularization of (\ref{eq:reg}) on the latent feature.}
\ENDFOR
\ENDWHILE
\end{algorithmic}
\end{algorithm}

By combining the feature and classifier module, the overall training scheme is referred as \textbf{subspace approximation classification (SAC)}. As summarized in Algorithm~\ref{alg:main} and visualized in Fig.\ref{fig:main_fig}, SAC takes a task from the LCL problem as input. Given a task $t_j$ in (\ref{eq:task_def}), the feature reduction module maps the inputs associated to $t_j$ into a $k$ dimension latent representation, where the label dependent eigenvectors are learned and stored vector bank. During inference, the feature module is again applied to the input and the output latent representation is then passed through the QDC classifier for obtaining final prediction. 

\begin{table*}
\caption{ Comparison with baselines under class incremental learning scenario.}
\centering
\begin{tabular}{|c||ccc||ccc|}
    \hline
     &  \multicolumn{3}{c||}{Datasets (Accuracy)} 
      & \multicolumn{3}{c|}{Datasets (Training Time (sec))}\\
    Methods & Mnist & KMnist & Fashion Mnist & Mnist & KMnist & Fashion Mnist\\
    \hline
    \hline
    DGR~\cite{dgr}  & 90.44$\pm$1.56 & 69.25$\pm$2.94  & 74.83$\pm$5.50 & 315.99$\pm$2.25 & 748.75$\pm$51.17 & 760.21$\pm$21.72\\
    DGR+distill~\cite{dgr,lwf}  & 92.31$\pm$0.74 & 64.42$\pm$1.12 & 76.03$\pm$4.12 & 314.12$\pm$12.79 & 819.52$\pm$14.52 & 800.81$\pm$3.69\\
    EWC~\cite{ewc}  & 20.45$\pm$1.15 & 19.54$\pm$0.12 & 19.97$\pm$0.02 & 398.86$\pm$11.04 & 719.89$\pm$21.95 & 697.24$\pm$53.39  \\
    Online EWC~\cite{onlineewc}& 20.69$\pm$1.53 & 19.54$\pm$0.12 & 19.97$\pm$0.03 & 371.87$\pm$12.35 & 665.04$\pm$3.40 & 692.49$\pm$29.20\\
    iCaRL~\cite{icarl}  & 93.24$\pm$0.70 & 70.83$\pm$2.78 & 79.61$\pm$0.79 & 200.16$\pm$9.83 & 468.38$\pm$4.98 & 466.60$\pm$11.09 \\
    LwF~\cite{lwf}  & 20.98$\pm$0.85 & 20.16$\pm$0.24 & 19.42$\pm$2.54 & 198.40$\pm$9.09 & 495.62$\pm$31.48 & 499.49$\pm$8.77\\
    RtF~\cite{rtf}  & 93.75$\pm$1.28 & 66.16$\pm$3.06 & 74.11$\pm$4.82 & 253.37$\pm$9.22 & 639.66$\pm$25.56 & 678.42$\pm$34.04 \\
    SI~\cite{si}  & 19.85$\pm$0.10 & 19.53$\pm$0.09 & 19.97$\pm$0.02  & 194.16$\pm$87.6 & 503.72$\pm$5.15 & 498.37$\pm$3.28\\
    CNDPM~\cite{cndpm}  & 93.54$\pm$0.13 & 74.35$\pm$1.4  &  44.62$\pm$2.1 & $>$ 3600 &  $>$ 3600 & $>$ 3600\\
    Saak~\cite{saak}& 95.21 &76.25 & 73.51& $>$ 3000& $>$ 3000 & $>$ 3000\\
    \hline
    \hline
    Ours & \textbf{95.59} & \textbf{77.35} & \textbf{80.32} & \textbf{5.90} & \textbf{5.72} & \textbf{5.48} \\
    \hline
\end{tabular}
\label{tab:main_full}
\end{table*}

\subsubsection{Efficient learning with regularization}
While SAC provides an efficient classification solution for the LCL problem, it can be problematic, because the inverse of $\hat{\Sigma}_c$ in (\ref{eq:pxy}) is an ill-defined problem numerically if it is close to a singular matrix. To solve the problem, the relationship between $\Sigma_c$ and $\hat{\Sigma}_c$ is revisited. By leveraging the latent representation $f(x)=\hat{Q}_c^Tx \in \mathcal{R}^k$ defined in Sec.~\ref{sec:saab_plus}, the covariance of the latent representation can be reformulated as
\begin{align}
\hat{\Sigma}_c &= Cov(f(x)) = \hat{Q}_c^T Cov(x) \hat{Q}_c \nonumber\\
&= \hat{Q}_c^T{\Sigma}_c\hat{Q}_c = \hat{Q}_c^T Q_c\Lambda_cQ_c^T \hat{Q}_c = \hat{\Lambda}_c.
\label{eq:relation_2sigma}
\end{align}
To avoid singularity, we proposed to add a regularization on $\hat{\Sigma}_i$ of (\ref{eq:pxy}) as 
\begin{equation}
    \hat{\Sigma}_c' =\hat{\Sigma}_c + \alpha*I = \hat{\Lambda}_c + \alpha*I, \quad \forall c\in\mathcal{Y} 
    \label{eq:reg}
\end{equation} 
when singular matrix occurs and this is referred as \textbf{regularized SAC (RSAC)}. 
The use of RSAC not only avoids the singularity, but also simplifies the computation of (\ref{eq:pxy}) as the inversion of $\hat{\Sigma}_c'$ is also a diagonal matrix composed of the eigenvalues from $\Lambda_c$. Note that RSAC does not require large number of parameter learning and is more computation and memory efficiency for the lightweight continuous learning applications.


\begin{figure}
\centering
\begin{tabular}{cccc}
\multicolumn{4}{c}{Mnist} \\
    \includegraphics[width=0.18\linewidth]{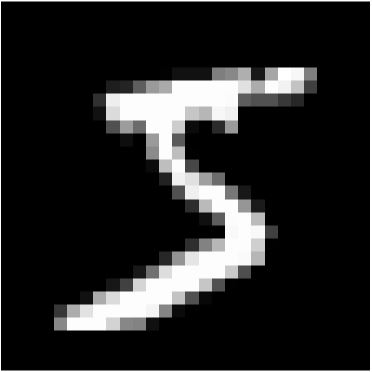} & 
    \includegraphics[width=0.18\linewidth]{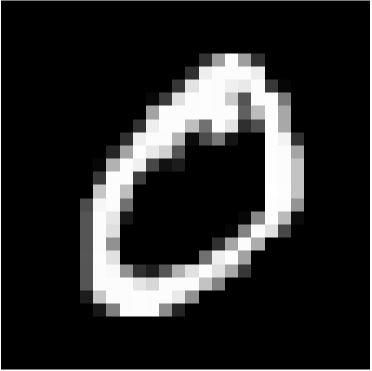} & 
    \includegraphics[width=0.18\linewidth]{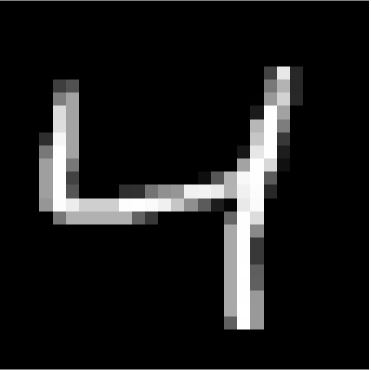} & 
    \includegraphics[width=0.18\linewidth]{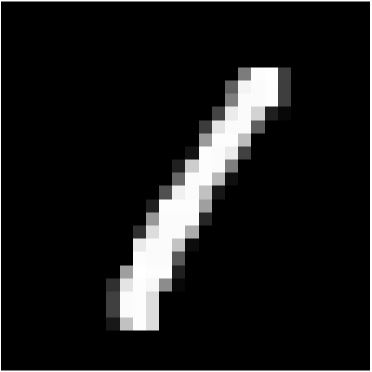}\\
    \multicolumn{4}{c}{KMnist} \\
    \includegraphics[width=0.18\linewidth]{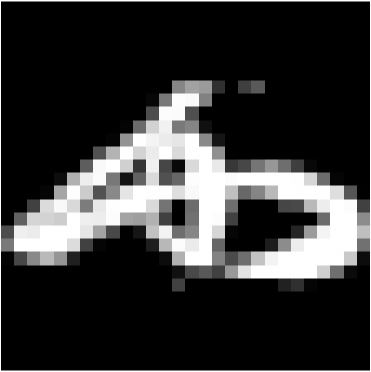} & 
    \includegraphics[width=0.18\linewidth]{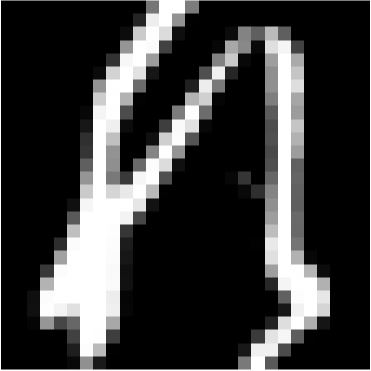} & 
    \includegraphics[width=0.18\linewidth]{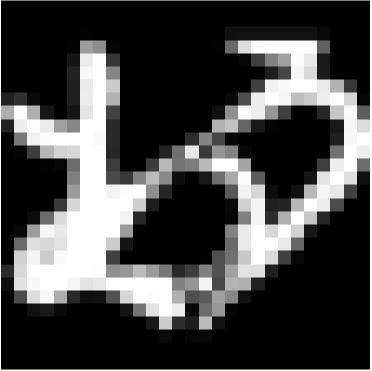} & 
    \includegraphics[width=0.18\linewidth]{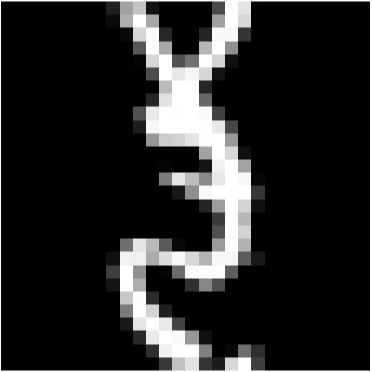}\\
    \multicolumn{4}{c}{FashionMnist} \\
    \includegraphics[width=0.18\linewidth]{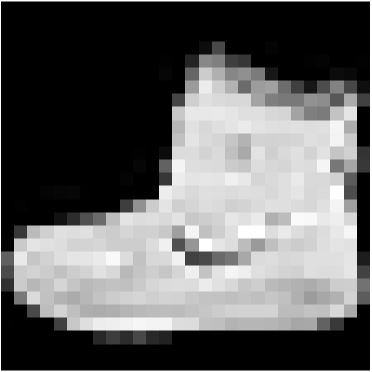} & 
    \includegraphics[width=0.18\linewidth]{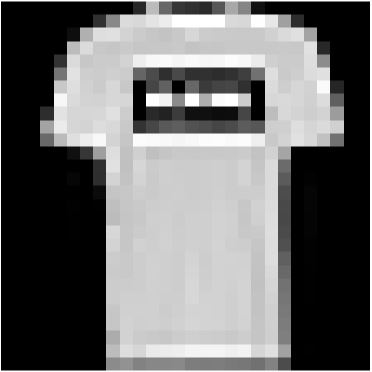} & 
    \includegraphics[width=0.18\linewidth]{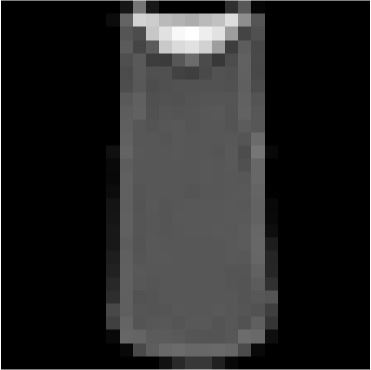} & 
    \includegraphics[width=0.18\linewidth]{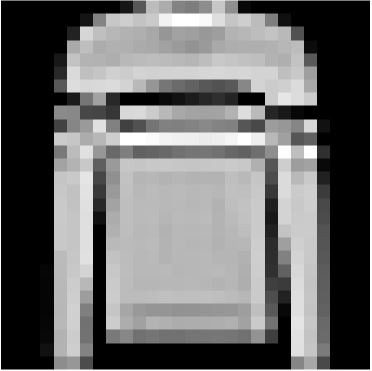}\\
    
\end{tabular}
\caption{ Visualization of the examples in 3 datasets.}
\label{fig:data_fig}
\end{figure}

\section{Experiment}

In this section,  Mnist~\cite{mnist}, KMnist~\cite{kmnist} and FashionMnist~\cite{fashionmnist} are evaluated in terms of averaged classification accuracy over all classes. Examples from all three datasets are visualized in Fig.~\ref{fig:data_fig}. For all three datasets, there are 10 classes, 60000 training images and 10000 testing images. Unlike KMnist and FashionMnist, Mnist has a slightly imbalance data distribution across classes.


Nine different continuous learning baselines are compared. The official code \footnote{\url{https://github.com/soochan-lee/CN-DPM}} of CNDPM~\cite{cndpm} is adopted and the public available code\footnote{\url{https://github.com/GMvandeVen/continual-learning}} for the rest of the baselines is used. Note that while the original architecture of CNDPM is used, the rest of the baselines are implemented based on multilayer perceptrons (MLP) as backbones. 
In addition, although the Saak architecture is not designed specifically for continuous learning scenario, it can be compared with slight modification of its official code\footnote{\url{https://github.com/davidsonic/Saak-Transform}}. 
To avoid the memory explosion, 
we implement three stages Saak transform with transformation sizes $8\times8$, $2\times2$ and $2\times2$, and use KLT transform and regularized QDC for classification 
in order to fit the continuous learning scenario. For a fair comparison, the number of feature selected is the same between RSAC and Saak.

For continuous learning settings, 5 tasks are considered by separating the dataset into pairs (i.e. grouping class (0/1), (2,3) etc.). The regularization hyperparmeter $\alpha$ of (\ref{eq:reg}) is set as 0.4 for all datasets.

\begin{figure}
\centering
\begin{tabular}{ccc}
    \includegraphics[width=1.00\linewidth]{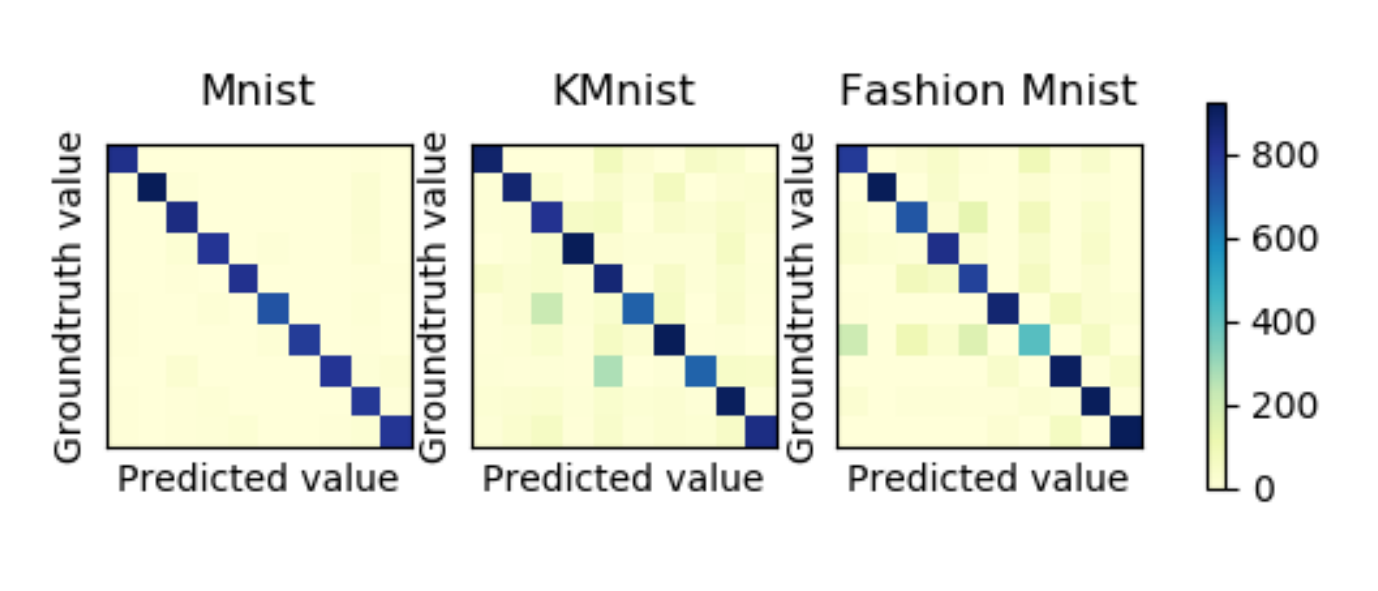}
\end{tabular}
\caption{ Confusion matrix of Mnist, KMnist, FashionMnist from the QDC classifier.}
\label{fig:confusion_mat}
\end{figure}

\subsection{Comparison with continuous learning baselines}
    

The continuous learning baselines are compared from the perspectives of classification accuracy and training time, as these 2 factors are critical to perform lightweight continuous learning (LCL). As shown in left of Table~\ref{tab:main_full}, the proposed framework outperforms the continuous learning baselines both in terms of accuracy and training time. The accuracy gain is more than $0.38\%$, $1.10\%$ and $0.71\%$ on Mnist, KMnist and FashionMnist respectively. The corresponding confusion matrix of the proposed architecture is visualized in Fig.~\ref{fig:confusion_mat}.

In addition to the accuracy gain, the training time of proposed framework is significantly smaller than those baselines implemented with deep network, as shown in right of Table~\ref{tab:main_full}. Such efficiency is attributed to the discard of backpropagation during training and the prevention of memory explosion. Note that most of the continuous learning baselines, besides CNDPM, are implemented with 4 layers MLP in order to provide more competitive baselines, but the proposed framework still beats those baselines with significant margin without sacrificing the accuracy. The results presented in Table~\ref{tab:main_full} demonstrate the efficiency and effectiveness of the proposed algorithm and suggest that the proposed framework is a more suitable solution for LCL problem.

\subsection{Ablation study}
In this section, the ablation study of the proposed framework is conducted by investigating the effect of power threshold $t$ of (\ref{eq:power_thres}) and the number of training data provided to the proposed framework.

\subsubsection{Effect of power threshold}
From Table~\ref{tab:ablation_k}, it can be observed that the accuracy saturates when the power threshold $t$ is around $0.95$ for all three datasets. The benefit of adding more eigenvectors is marginal when $t>0.95$. Moreover, when adding eigenvectors associated with small eigenvalues, the computation of QDC (\ref{eq:map_log}) will often lead to numerical error as the covariance matrix is not invertible. The proposed regularization of (\ref{eq:reg}) can avoid such ill-defined inverse matrix with minor drop of accuracy.

Furthermore, while the input has 784 dimension, it is not necessary to store all 784 eigenvectors in the vector bank. Instead, the best accuracy reported in Table~\ref{tab:ablation_k} shows that storing less than 200 eigenvectors per class is sufficient to achieve competitive performance. With such memory efficiency, it allows the edge device to classify more data from more classes under same amount of memory budget.

\begin{table}
\caption{Different power thresholds $t$ of (\ref{eq:power_thres}) are explored. Each threshold corresponds to a specific $k$ value in (\ref{eq:power_thres}). The $k$ value associated to the best accuracy  are reported. }
\label{tab:ablation_k}
\centering
\begin{tabular}{|ccccccc|}
    \hline
    Power threshold $t$  & \multicolumn{2}{c}{Mnist} & \multicolumn{2}{c}{KMnist} & \multicolumn{2}{c|}{Fashion Mnist}\\
    &$k$ & acc & $k$ & acc & $k$ & acc\\
    \hline
    0.8 & 31 & 67.75 & 64 & 61.30 & 26 & 64.90\\
    0.9 & 68 & 93.22 & 126 & 76.16 & 77 & 73.98\\
    0.95 & 121 & 95.41 & 211 & 77.13 & 156 & 79.74\\
    0.96 & 141 & 95.43 & 243 & 76.87 & 185 & 80.25\\
    0.97 & 168 & 95.43 & 285 & 74.84 & 224 & 73.56\\
    0.98 & 206 & 91.66 & 346 & 75.09 & 278 & 73.95\\
    \hline
    \hline
    Best & 150 & 95.59 & 192 & 77.35 & 183 & 80.32  \\
    \hline
\end{tabular}
\end{table}

\begin{table}
\caption{Comparison with Saak under data incremental learning scenario.}
\centering
\begin{tabular}{|cccc|}
    \hline
    Methods/Datasets & Mnist & KMnist & Fashion Mnist \\
    \hline
    Saak~\cite{saak} & 95.59 & 77.00 & 78.15 \\
    \hline
    Ours & \textbf{95.92} & \textbf{80.48} & \textbf{80.24}  \\
    \hline
\end{tabular}
\label{tab:main_incremental}
\end{table}

\subsubsection{Effect of dataset size}
\begin{figure}
\centering
\begin{tabular}{c}
    \includegraphics[width=0.8\linewidth]{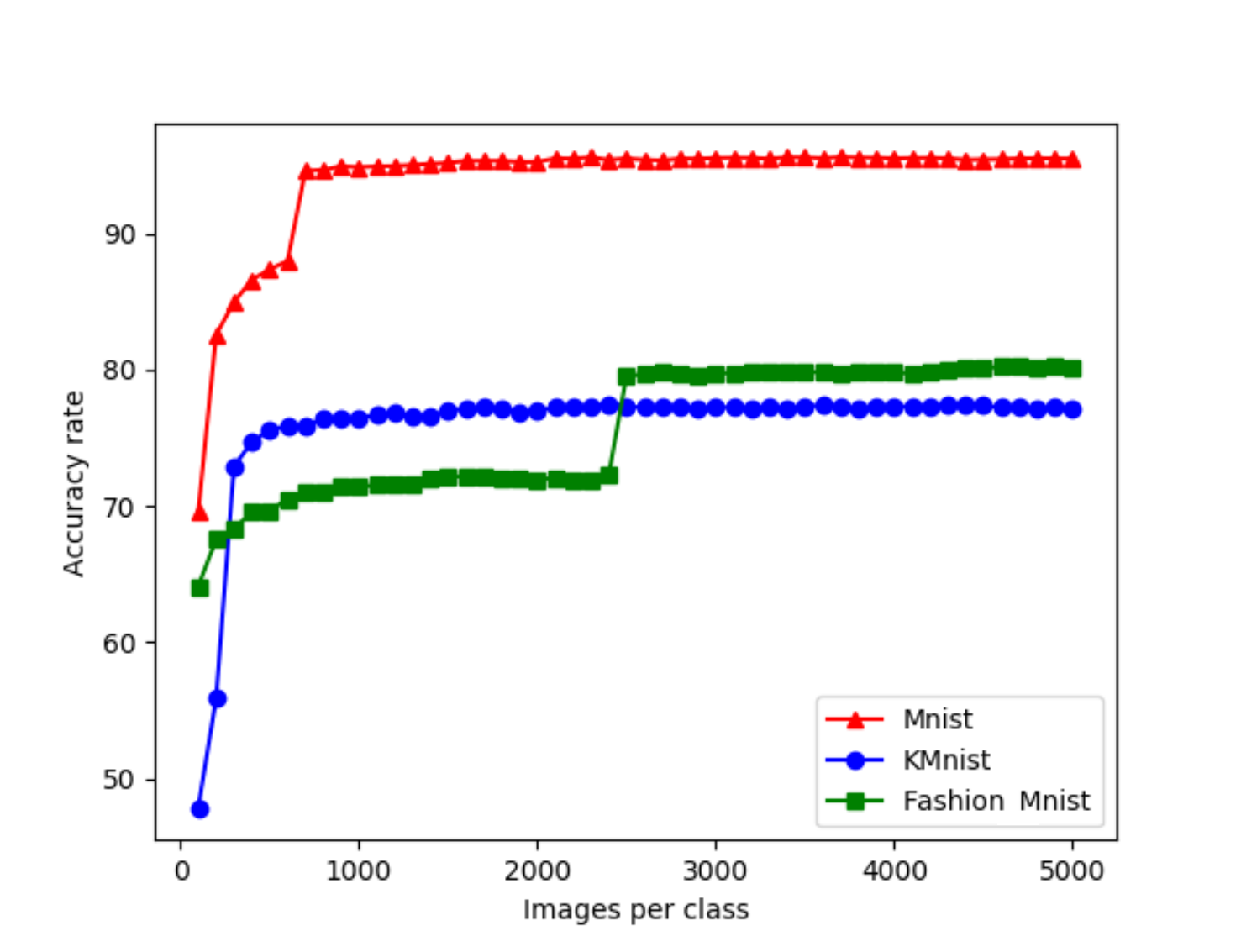}
\end{tabular}
\caption{Ablation study on the number of data needed for training a good LCL classifier. The accuracy saturates after the use of $15\%$, $10\%$ and $50\%$ images for Mnist, KMnist, Fashion Mnist, respectively.}
\label{fig:ablation_data_size}
\end{figure}

It is known that the deep neural network requires a large number of labeled data for learning a good classifier~\cite{datasize}. The dependency of large number of labeled data is investigated on the proposed method, as shown in Fig.~\ref{fig:ablation_data_size}. Unlike the heavy dependency of neural network based methods, the proposed framework is data efficiency as the accuracy remains fairly stable when more than 500 (800) images per class are used, which only requires less than $15\%$, $10\%$ of the available training data in Mnist (KMnist). For a more challenging dataset Fashion Mnist, merely $50\%$ of the available training data is needed to achieve competitive results.
This indicates that the proposed framework is more suitable in the application where training data is scarce.

\subsection{Data incremental continuous learning}
Inspired by the observation in the ablation study that the proposed method is less sensitive to the number of training data, the proposed method is further evaluated under a novel continuous learning scenario, where \textbf{data from same class} does not stream in the same time stamp. We refer this as \textit{data incremental continuous learning}. Note that this is different from the typical continuous learning settings discussed in Sec.~\ref{sec:setting}, where data from same class always arrive at the same time stamp. Since most prior works in continuous learning literature does not fit into this scenario, the only comparable baseline is SaaK~\cite{saak}. As shown in Table~\ref{tab:main_incremental}, the proposed method beats SaaK for all 3 datasets. Moreover, the gain increases from $0.33\%$ of simple dataset (i.e. Mnist) to $2.09\%$ of a more challenging dataset (i.e Fashion Mnist).

\section{Conclusion} 
In this paper, the underlying disadvantages of current continuous learning algorithms are discussed, including the slow training time, memory inefficiency and dependency on large number of training data. All factors harm the usability of continuous learning algorithms in the real world application, where the streaming data is limited and the training time is critical. As the result, the importance of lightweight continuous learning problem is investigated with the proposed algorithm regularized subspace approximation classifier (RSAC). RSAC inherits the advantages of previous lightweight transformation with new architecture to match continuous learning scenario. RSAC also consists a quadratic discriminant classifier (QDC) with addition regularization to prevent ill-defined condition. Moreover, the elaborate design of RSAC reduces the cost of memory storage and enables the classification to be perform in an efficient manner. Extensive experiments demonstrate the performance as well as training speed of the proposed framework. We hope this work will inspire the focus of lightweight continuous learning problem in the literature.


    


{\small
\bibliographystyle{unsrt}
\bibliography{ref}
}

\end{document}